\documentclass[runningheads]{llncs}
\usepackage[T1]{fontenc}
\usepackage{graphicx}
\usepackage{amsmath,amssymb}
\usepackage{booktabs}
\usepackage{pifont}
\usepackage{multirow}
\usepackage{makecell}
\usepackage{algorithm}
\usepackage{algpseudocode}

\usepackage[frozencache=true,cachedir=minted-cache]{minted}

\newcommand{\hide}[1]{}

\usepackage{xcolor}
\usepackage{amssymb}

\colorlet{nfcolor}{teal}
\definecolor{tmcolor}{rgb}{0.2,0.6,0.6}
\definecolor{todocolor}{rgb}{0.749,0.212,0.047}
\definecolor{changedcolor}{rgb}{0.42,0.27,0.57}

\begin{document}
\definecolor{ourgreen}{RGB}{152,190,109} 
\definecolor{ourred}{RGB}{195,117,131} 
\title{WikiCoder: Learning to Write Knowledge-Powered Code}
%\title{Contribution Title\thanks{Supported by organization x.}}
%
%\titlerunning{Abbreviated paper title}
% If the paper title is too long for the running head, you can set
% an abbreviated paper title here
%
\author{Théo Matricon\inst{1}\orcidID{0000-0002-5043-3221} \and \\
Nathana{\"e}l Fijalkow\inst{1,2}\orcidID{0000-0002-6576-4680} \and \\
Ga{\"e}tan Margueritte\inst{1}\orcidID{0009-0006-1742-4688}}

\authorrunning{T. Matricon, N. Fijalkow, G. Margueritte}
% First names are abbreviated in the running head.
% If there are more than two authors, 'et al.' is used.

\institute{CNRS, LaBRI, University of Bordeaux, France \email{\{theomatricon,gamargueritte\}@gmail.com}\and
MIMUW, University of Warsaw, Poland
\email{nathanael.fijalkow@gmail.com}}

\maketitle              % typeset the header of the contribution

\begin{abstract}
We tackle the problem of automatic generation of computer programs from a few pairs of input-output examples. The starting point of this work is the observation that in many applications a solution program must use external knowledge not present in the examples: we call such programs knowledge-powered since they can refer to information collected from a knowledge graph such as Wikipedia.
This paper makes a first step towards knowledge-powered program synthesis. We present WikiCoder, a system building upon state of the art machine-learned program synthesizers and integrating knowledge graphs. We evaluate it to show its wide applicability over different domains and discuss its limitations. WikiCoder solves tasks that no program synthesizers were able to solve before thanks to the use of knowledge graphs, while integrating with recent developments in the field to operate at scale.

\end{abstract}

\section{Introduction}
Task automation is becoming increasingly important in the digital world we live in, yet writing software is still accessible only to a small share of the population. Program synthesis seeks to make coding more reliable and accessible by developing methods for automatically generating code~\cite{Gulwani2017}. For example, the FlashFill system~\cite{Gulwani2011} in Microsoft Excel makes coding more accessible by allowing nontechnical users to synthesize spreadsheet programs by giving input-output examples, and TF-coder~\cite{ShiBS22} assists developers with writing TensorFlow code for manipulating tensors.

Very impressive results have been achieved in the past five years employing machine learning methods to empower program synthesis.
Recent works have explored engineering neural networks for guiding program search~\cite{Balog2017,Devlin2017,Lee2018,Zhang2018,Polosukhin2018,Kalyan2018,ZoharW18,chen2018execution,EllisWNSMHCST21,Fijalkow2022ScalingNP} effectively by training the network to act as a language model over source code.
The most resounding success of this line of work is OpenAI's Codex system~\cite{abs-2107-03374} powering Github's Copilot and based on very large language models.
However, because it works at a purely syntactic level, program synthesis fails in a number of applications.

Let us consider as an example the following task:
\begin{minted}{python}
f("17", "United States") = "17 USD"
f("42", "France") = "42 EUR"
\end{minted}
The task is specified in the programming by example setting: the goal is to construct a function \texttt{f} mapping inputs to their corresponding outputs.
Solving this task requires understanding that the second inputs are countries and mapping them to their currencies.
This piece of information is not present in the examples and therefore no program synthesis tool can solve that task without relying on external information.
In other words, a solution program must be \textit{knowledge-powered}!
An example knowledge-powered program yielding a solution to the task above is given below in a Python-like syntax:
\begin{minted}{python}
def f(x,y):
    return x + " " + CurrencyOf(y)
\end{minted}
It uses a function \texttt{CurrencyOf} obtained from an external source of knowledge.

Knowledge-powered program synthesis extends classical program synthesis by targetting knowledge-powered programs.
The challenge of combining syntactical manipulations performed in program synthesis with semantical information was recently set out by~\cite{VerbruggenLG21}. They discuss a number of applications: string manipulations, code refactoring, and string profiling, and construct an algorithm based on very large language models (see related work section).

Our approach is different: the methodology we develop relies on knowledge graphs, which are very large structured databases organising knowledge using ontologies to allow for efficient and precise browsing and reasoning. There is a growing number of publicly available knowledge graphs, for instance Wikidata.org, which includes and structures Wikipedia data, and Yago~\cite{TanonWS20,HoffartSBW13}, based on Wikidata and schema.org. We refer to~\cite{2021Hogan} for a recent textbook and to~\cite{Hogan22} for an excellent survey on knowledge graphs and their applications, and to Figure~\ref{fig:knowledge_graph} for an illustration.

The recent successes of both program synthesis and knowledge graphs suggest that the time is ripe to combine them into the knowledge-powered program synthesis research objective.

\paragraph*{Our contributions:}
\begin{itemize}
	\item We introduce knowledge-powered program synthesis, which extends program synthesis by allowing programs to refer to external information collected from a knowledge graph.
	\item We identify a number of milestones for knowledge-powered program synthesis and propose a human-generated and publicly available dataset of 46 tasks to evaluate future progress on this research objective.
	\item We construct an algorithm combining state of the art machine learned program synthesizers with queries to a knowledge graph, which can be deployed on any knowledge graph.
	\item We implement a general-purpose knowledge program synthesis tool WikiCoder and evaluate it on various domains.
	WikiCoder solves tasks previously unsolvable by any program synthesis tool, while still operating at scale by integrating state of the art techniques from program synthesis.
\end{itemize}

%%%%%%%%%%%%%%%%%%%%%%%%%%%
\section{Knowledge-Powered Programming by Example}

\subsection{Objectives}
Programming by example, or inductive synthesis, is the following problem: given a few examples, construct a program satisfying these examples. This particular setting in program synthesis, where the user gives a very partial specification, has been extremely useful and successful for automating tasks for end users, the prime example being FlashFill for performing string transformations in Excel~\cite{Gulwani2011}.

When considering knowledge-powered programming by example, we do not change the problem, only the solution: instead of classical programs performing syntactic manipulations, we include knowledge-powered programs. To illustrate the difference, let us consider the following two tasks.
\begin{minted}{python}
f("Paris") = "I love P"
f("Berlin") = "I love B"

g("Donald Knuth") = "DK is American"
g("Ada Lovelace") = "AL is English"
\end{minted} 
The first is a classical program synthesis task in the sense that it is purely syntactical, it can be solved with a two-line program concatenating ``I love '' with the first letter of the input.
On the other hand, the second requires some knowledge about the input individuals, here their nationality: one needs a knowledge-powered program to solve this task.
Since almost all program synthesis tools perform only syntactical manipulations of the examples (we refer to the related work section for an in-depth discussion), they cannot solve the second task.

Knowledge-powered programming by example goes much beyond query answering: the goal is not to answer a particular query, but to produce a program able to answer that query for any input. This is computationally and conceptually a much more difficult problem.

\begin{figure}[ht]
   \centering
   \includegraphics[width=0.8\textwidth]{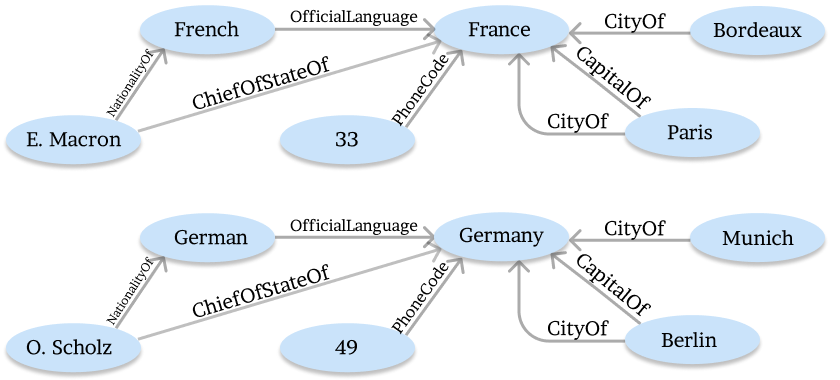}
   \caption{Illustration of part of a knowledge graph.}
   \label{fig:knowledge_graph}
\end{figure}

For concreteness we introduce some terminology about knowledge graphs, and refer to Figure~\ref{fig:knowledge_graph} for an illustration.
Nodes are called entities, and edges are labelled by a relation. Entities are arranged into classes: ``E. Macron'' belongs to the class of people, and ``France'' to the class of countries. The classes and relations are constrained by ontologies, which define which relations can hold between entities.
The de facto standard for querying knowledge graphs is through \texttt{SPARQL} queries, which is a very powerful and versatile query language.

\subsection{Milestones}\label{sec:milestones}
We identify three independent ways in which semantical information can be used.
They correspond to different stages for solving a task:
\begin{itemize}
	\item \textit{preprocessing:} the first step is to extract entities from the examples;
	\item \textit{relating:} the second step is to relate entities in the knowledge graph;
	\item \textit{postprocessing:} the third step is to process the information found in the knowledge graph.
\end{itemize}

\paragraph*{How much preprocessing to extract entities?}
Let us consider two tasks.
\begin{minted}{python}
f("Aix, Paris, Bordeaux") = "Paris"
f("Hamburg, Berlin, Munich") = "Berlin"

g("President Obama") = "Obama"
g("Prime Minister de Pfeffel Johnson") = "de Pfeffel Johnson"
\end{minted}
In the function \texttt{f} the goal is to extract the second word as separated by commas: this is a purely syntactic operation.
In the function \texttt{g} we need to remove the job title from the input: this requires semantical knowledge, for instance it is not enough to use neither the second nor the last word.

\paragraph*{How complicated is the relationship between entities?}
We examine two more tasks.
\begin{minted}{python}
f("Paris") = "France is beautiful"
f("Berlin") = "Germany is beautiful"

g("Paris") = "Phone country code: 33"
g("Berlin") = "Phone country code: 49"
\end{minted}
The function \texttt{f} relates two entities: a city and a country. One can expect that the knowledge graph includes the relation \texttt{CapitalOf}, which induces a labelled edge between ``Paris'' and  ``France'' as well as ``Berlin'' and ``Germany'' (as in Figure~\ref{fig:knowledge_graph}). Note that it could also be the relation \texttt{CityOf}.
More complex, the function \texttt{g} requires crossing information: indeed to connect a city to its country code, it is probably required 
to compose two relations: \texttt{CapitalOf} and \texttt{PhoneCode}. In other words, the entities are related by a path of length $2$ in the knowledge graph, that we write \texttt{CapitalOf-PhoneCode}.
More generally, the length of the path relating the entities is a measure of complexity of a task.

\paragraph*{How much postprocessing on external knowledge?}
Let us look again at two tasks.
\begin{minted}{python}
f("Paris") = "Country's first letter: F"
f("Berlin") = "Country's first letter: G"

g("President Obama") = "B@r@ck"
g("Prime Minister Johnson") = "B0r1s"
\end{minted}
For the function \texttt{f}, the difficulty lies in finding out that the external knowledge to be search for relates ``Paris'' to ``France'' and ``Berlin'' to ``Germany'', and then to return only the first letter of the result. 
Similarly, for \texttt{g}, before applying a leet translation we need to retrieve as external knowledge the first name.
In both cases the difficulty is that the intermediate knowledge is not present in the examples.

\vskip1em
The three steps can be necessary, the most complex tasks involve at the same time subtle preprocessing to extract entities, complicated relationships between entities, and significant postprocessing on external knowledge.

\subsection{Motivating Examples}\label{sec:applications}
We illustrate the disruptive power of knowledge-powered program synthesis in three application domains.

\paragraph*{General knowledge}
The first example, which we use throughout the paper for illustrations, in the dataset and in the experiments, is to use the knowledge graph to obtain general facts about the world, such as geography, movies, people. Wikidata and Yago are natural knowledge graphs candidates for this setting. 
This domain is heavily used for query answering: combining it with program synthesis brings it to another level, since programs generalize to any input.

\paragraph*{Grammar exercises}
The second example, inspired from~\cite{VerbruggenLG21}, is about language learning: tasks are grammar exercises, where the goal is to write a grammatically correct sentence. Here the knowledge graph includes grammatical forms and their connections, such as verbs and their different conjugated forms, pronouns, adjectives, and so on.
Generating programs for solving exercises opens several perspectives, including generating new exercises as well as solving them automatically.

\paragraph*{Advanced database queries}
In the third example knowledge-powered program synthesis becomes a powerful querying engine. This scenario has been heavily investigated for SQL queries~\cite{WangCB17,ZhouBCW22}, but only at a syntactic level.
Being able to rely on the semantic properties of the data opens a number of possibilities, let us illustrate them on an example scenario.
The knowledge graph is owned and built by a company, it contains immutable data about products.
The database contains customer data.
Crossing semantical information between the database being queried and the knowledge graph allows the user to generate complex queries to extract more information from the database including for instance complex statistics.

%%%%%%%%%%%%%%%%%%%%%%%%%%%
\section{WikiCoder}

\begin{figure}[t]
   \centering
   \includegraphics[width=0.8\textwidth]{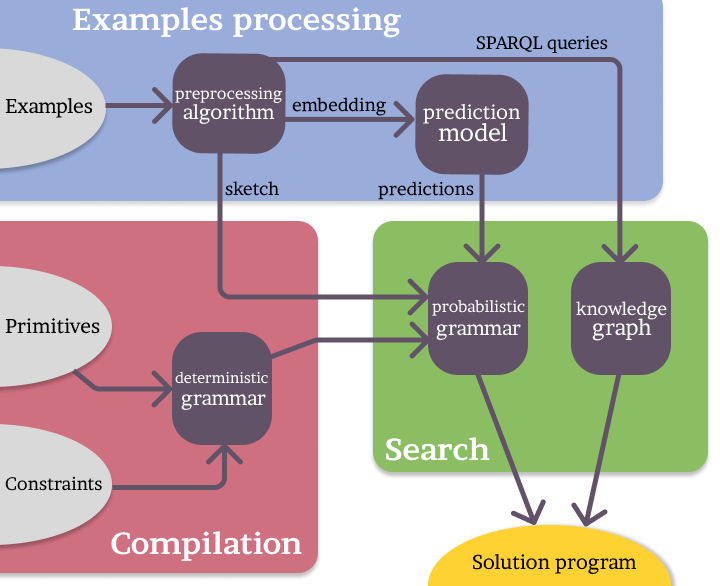}
   \caption{WikiCoder pipeline.}
   \label{fig:pipeline}
\end{figure}

WikiCoder is a general-purpose grammar-based program synthesis tool, it was developed in Python integrating state of the art program synthesis techniques. It is publicly available on GitHub\footnote{\url{https://github.com/nathanael-fijalkow/ProgSynth/tree/WikiCoder}}.
The main functionality of WikiCoder is to solve programming by example tasks: the user inputs a few examples and WikiCoder synthesizes a knowledge-powered program satisfying the examples.
The programming language is specified as a domain specific language (DSL), designed to solve a common set of tasks.
WikiCoder supports a number of classical DSLs, including towers building~\cite{EllisWNSMHCST21}, list integer manipulations~\cite{Balog2017}, regular expressions, and string manipulation tasks à la FlashFill~\cite{Gulwani2011}: our experiments report on the latter DSL.
Following recent advances in machine learned program synthesizers~\cite{EllisWNSMHCST21,Fijalkow2022ScalingNP}, WikiCoder is divided into three components: compilation, examples processing, and search, as illustrated in Figure~\ref{fig:pipeline} and discussed in the next three sections.
This is very different from Codex's architecture, which is based on auto-regressive very large models directly generating code.

%%%%%%%%%%%%%%%%%%%%%%%%%%%
\subsection{Compilation}
The DSL is specified as a list of primitives together with their types and semantics.
The compilation phase consists in obtaining an efficient representation of the set of programs in the form of a context-free grammar.
The right way to look at the grammar is as a generating model: it generates code. 
Thanks to the expressivity of context-free grammars, many syntactic properties can be ensured: primarily and most importantly, all generated programs are correctly typed.

The DSL we use in our experiments is tailored for string manipulation tasks à la Flashfill. 
For the sake of presentation we slightly simplify it. 
We use two primitive types: \verb+STRING+ and \verb+REGEXP+, and one type constructor \verb+Arrow+.
We list the primitives below, they have the expected semantics.

\begin{minted}{python}
    $       : REGEXP                  # end of string
    .       : REGEXP                  # all
    [^_]+   : Arrow(STRING, REGEXP)   # all except X
    [^_]+$  : Arrow(STRING, REGEXP)   # all except X at the end
    compose : Arrow(REGEXP, Arrow(REGEXP, REGEXP))

    concat  : Arrow(STRING, Arrow(STRING, STRING))
    match   : Arrow(STRING, Arrow(REGEXP, STRING))

    # concat_if: concat if the second argument (constant) 
    #            is not present in the first argument
    concat_if    : Arrow(STRING, Arrow(STRING, STRING))  
    # split_fst: split using regexp, returns first result
    split_fst  : Arrow(STRING, Arrow(REGEXP, STRING))
    # split_snd: split using regexp, returns second result
    split_snd : Arrow(STRING, Arrow(REGEXP, STRING))
\end{minted} 

In the implementation we use two more primitive types: \verb+CONSANT_IN+ and \verb+CONSTANT_OUT+, which correspond to constants in the inputs and in the output. This is only for improving performances, it does not increase expressivity. Some primitives are duplicated to use the two new primitive types.

%%%%%%%%%%%%%%%%%%%%%%%%%%%
\subsection{Examples processing}
A preprocessing algorithm produces from the examples three pieces of information: a sketch, which is a decomposition of the current task into subtasks, a set of \texttt{SPARQL} queries for the knowledge graph, and an embedding of the examples for the prediction model.

\begin{figure}[ht]
   \centering
   \includegraphics[width=0.8\textwidth]{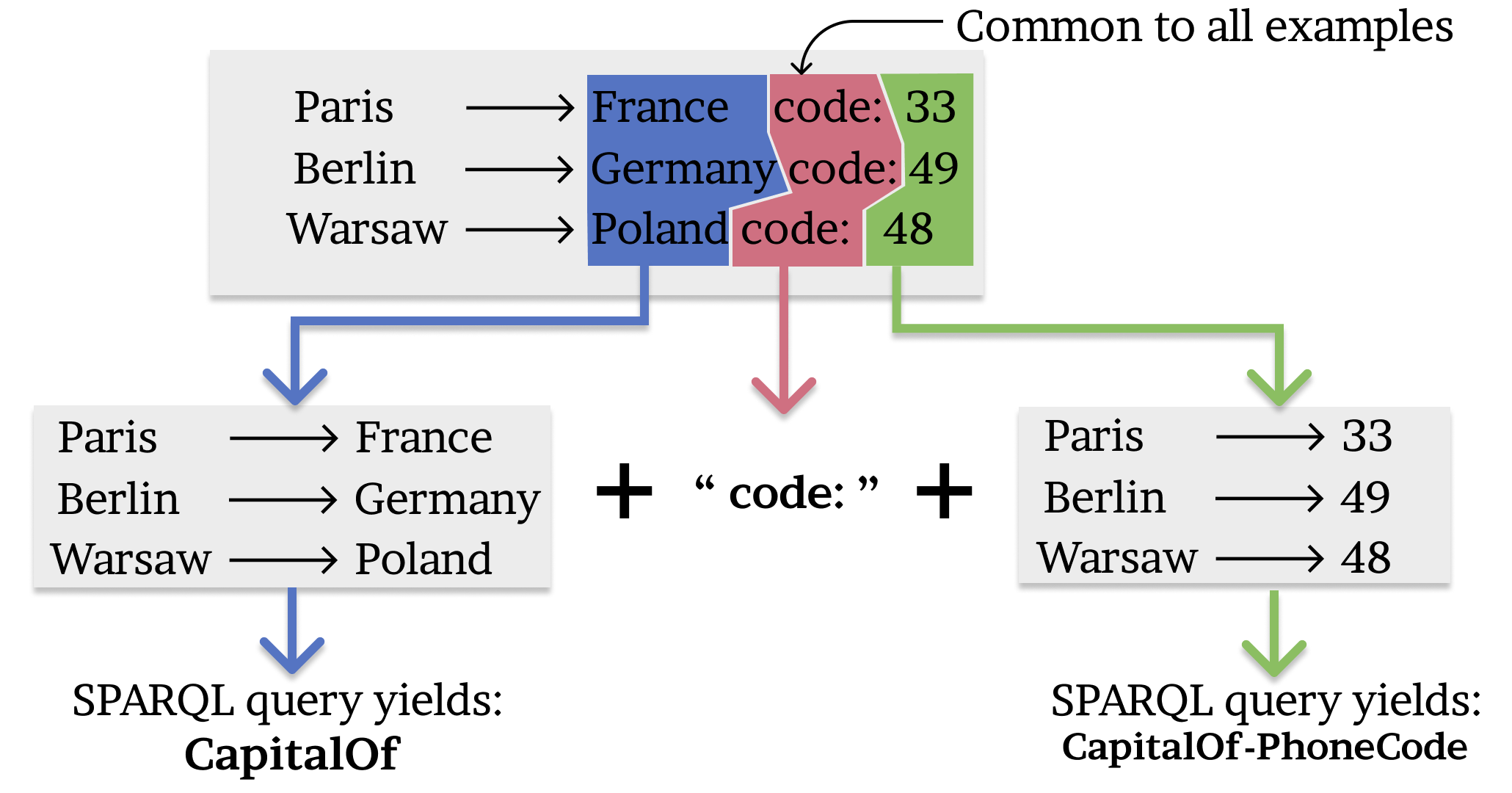}
   \caption{Example of the preprocessing algorithm.}
   \label{fig:preprocessing}
\end{figure}

\paragraph*{Preprocessing algorithm}
Figure~\ref{fig:preprocessing} illustrates the preprocessing algorithm in action.
The high level idea is that the algorithm is looking for a shared pattern across the different examples.
In the task above, `` code: '' is shared by all examples, hence it is extracted out.
A naive implementation is to look for the largest common factor between the strings, and proceed recursively on the left and on the right.
The process is illustrated with Algorithm~\ref{alg:preprocessing}, which produces a list of constants from a list of strings. This procedure is applied to both the inputs and outputs independently to create sketches.
Thanks to these constants the task is split into subtasks as illustrated in Figure~\ref{fig:preprocessing} where we have split the task into two subtasks: \texttt{CapitalOf} and \texttt{CapitalOf-PhoneCode}.
To solve each subtask we query the knowledge graph with the inputs and outputs.
If no path is found, we run a regular -- syntactical -- program synthesis algorithm.

\begin{algorithm}[ht]
   \caption{Constant extraction}\label{alg:preprocessing}
   \begin{algorithmic}
      \Procedure {GetConstants}{$S = (S_k)_{k \in [1,n]}$ : strings}
      \If{there is an empty string in $S$} 
         \State \textbf{return} empty list
      \EndIf
      \State $\text{factor} \gets$ longest common factor among all strings in $S$
      \If{$\text{len}(\text{factor}) \le 2$}
	      \State \textbf{return} empty list
	  \Else
	      \State $S_{\text{left}} \gets$ prefix of factor in $S$ 
	      \State $L_{\text{left}} \gets$ \textsc{GetConstants}$(S_{\text{left}})$
	      \State $S_{\text{right}} \gets$ suffix of factor in $S$ 
	      \State $L_{\text{right}} \gets$ \textsc{GetConstants}$(S_{\text{right}})$
          \State \textbf{return} $L_{\text{left}} + \text{factor} + L_{\text{right}}$
      \EndIf
      \EndProcedure
   \end{algorithmic}
\end{algorithm}

\paragraph*{Generated \texttt{SPARQL} queries}
The \texttt{SPARQL} queries are generated from the examples after preprocessing.
Once we have the constants with Algorithm~\ref{alg:preprocessing} of the inputs and outputs, we can split the inputs and outputs in constant parts and non-constant parts, only the non-consant parts are relevant.
For each non-constant part in the outputs, we generate queries from the non constant part of the inputs which should map to this non-constant part of the output. 
Since relations may be complex, that is ``Paris'' is at distance 1 from ``France'' but ``33'' is two relations away from ``Paris'', we generate \texttt{SPARQL} queries for increasing distances up to a fixed upper bound.
Here is the query at distance 2, that we execute for the example in Figure~\ref{fig:preprocessing} with \texttt{CapitalOf-PhoneCode}:
\begin{verbatim}
   PREFIX w: <https://en.wikipedia.org/wiki/>
   SELECT ?p0 ?p1 WHERE {
      w:Paris ?p0 ?o_1_0 .
      ?o_1_0 ?p1 w:33 .
      w:Berlin ?p0 ?o_2_0 .
      ?o_2_0 ?p1 w:49 .
      w:Warsaw ?p0 ?o_3_0 .
      ?o_3_0 ?p1 w:48 .
   }
\end{verbatim}
Notice that intermediary entities make an apparition in order to accommodate for longer path lengths.
The output of the above query would consist of two paths: \texttt{CityOf-PhoneCode} and \texttt{CapitalOf-PhoneCode}.

As disambiguation strategy (inspired by~\cite{ZhengSCLW22}) we choose the path with the least number of hits across all examples, called the least ambiguous path. 
In this example there is no preferred path since both paths lead to a single entity for each example.
%One could argue that \texttt{CityOf-PhoneCode} should be preferred since there are more starting entities than \texttt{CapitalOf-PhoneCodeOf}, thus it is more likely that we find such a path from our starting entity. 
%We did not implement such a mechanism but this could be done to select one path among the least ambiguous ones.
As an example of this disambiguation strategy, let us consider paths from ``33'' to ``Paris'': there are two paths, 
\texttt{PhoneCodeOf-Capital} and \texttt{PhoneCodeOf-City}.
The least ambiguous path is \texttt{PhoneCodeOf-Capital} since \texttt{PhoneCodeOf-City} leads to all cities of the country.

To find the least ambiguous path, we need to count the number of hits, which is done using more \texttt{SPARQL} queries.
Here is a sample query to get all entities at the end of the path \texttt{CapitalOf-PhoneCode} from the starting entity ``Paris'':
\begin{verbatim}
   PREFIX w: <https://en.wikipedia.org/wiki/>
   SELECT ?dst WHERE {
      w:Paris w:CapitalOf ?e0 .
      ?e0 w:PhoneCode ?dst .
   }
\end{verbatim}
We count the number of results for all examples to get the number of hits for a path.

\begin{figure}[ht]
   \centering
   \includegraphics[width=\textwidth]{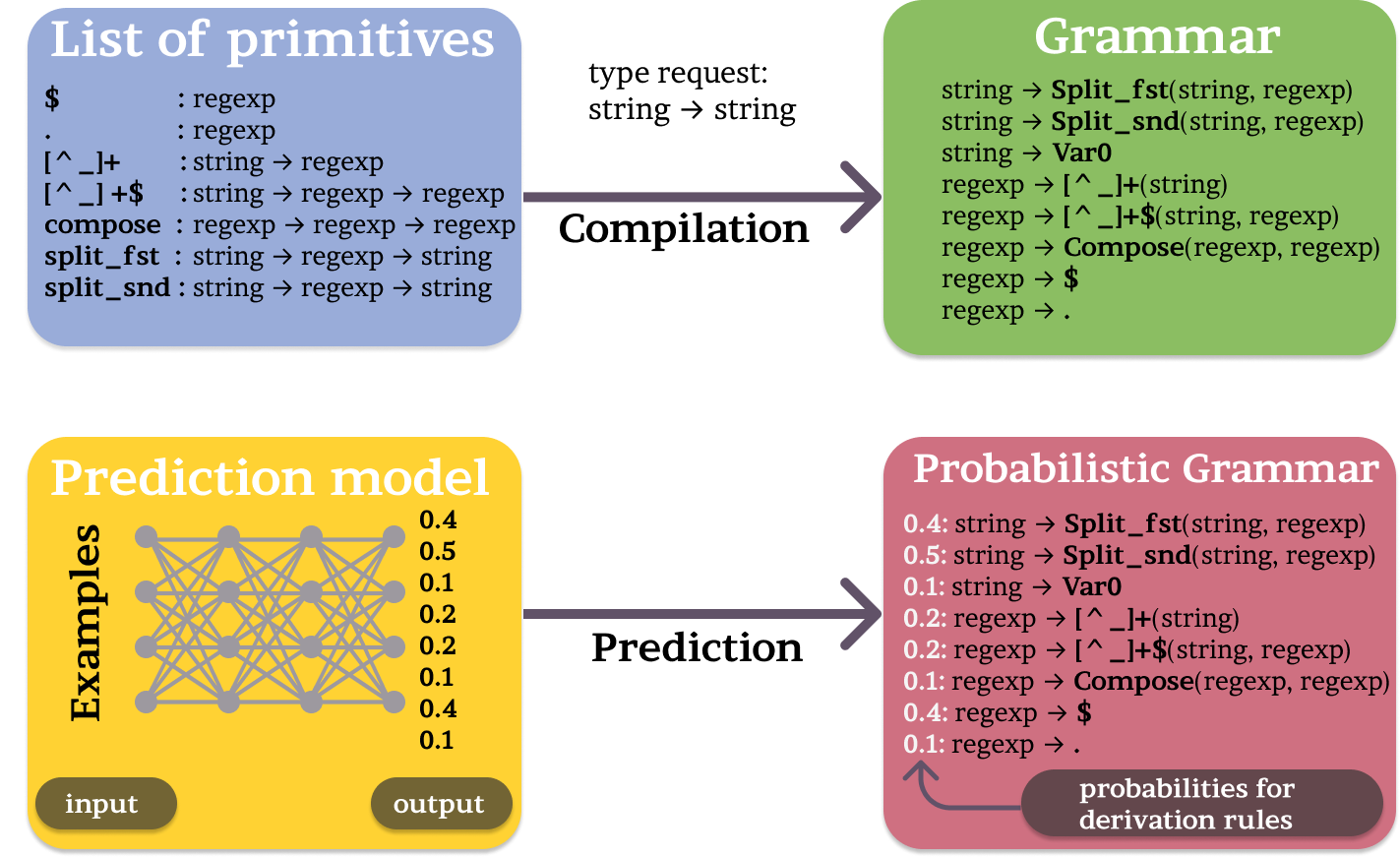}
   \caption{Illustration of the prediction model.}
   \label{fig:predictions}
\end{figure}

\paragraph*{Prediction model}
Efficient search is crucial in program synthesis, because the search space combinatorially explodes. We, and we believe the community more broadly, see learning as the right way of addressing this challenge.
Following~\cite{EllisWNSMHCST21,Fijalkow2022ScalingNP}, we build a prediction model in the form of a neural network: it reads embedding of the examples and outputs probabilities on the derivation rules of the grammar representing all programs. 
This transforms the context-free grammar representing programs in a probabilistic context-free grammar, which is a stochastic process generating programs. In other words, it defines a probabilistic distribution over programs, and the prediction model is trained to maximise the probability that a solution program is generated.

This prediction process is illustrated in Figure~\ref{fig:predictions}.
This effectively leverages the prediction power of neural networks without sacrificing correctness: the prediction model biases the search towards most likely programs but does not remove part of the search tree, hence -- theoretically -- a solution will be found if there exists one.

%%%%%%%%%%%%%%%%%%%%%%%%%%%
\subsection{Search}
The prediction model assigns to each candidate program a likelihood of being a solution to the task at hand. WikiCoder uses the HeapSearch algorithm, which is a bottom-up enumerative search algorithm~\cite{Fijalkow2022ScalingNP} outputting programs following the likelihood order of the prediction model. It has been shown superior to other enumeration methods (such as A* or Beam search) and easily deployed across parallel compute environments.
The candidate programs use the results of the \texttt{SPARQL} queries on the knowledge graph to be run on the examples.

%%%%%%%%%%%%%%%%%%%%%%%%%%%
\section{Evaluation}
We perform experiments to answer the following questions:
\begin{enumerate}
	\item[(Q1)] Which milestones as described in Section~\ref{sec:milestones} can be achieved with our algorithm?
	\item[(Q2)] How does WikiCoder compare to GPT-3 and Codex, the algorithm powering Copilot based on very large language models?
	\item[(Q3)] Can a knowledge-powered program synthesis tool operate at scale for classical purely syntactic tasks?
\end{enumerate}

\subsection{Environment}
The dataset along with the code for the experiments are available publicly on GitHub\footnote{\url{https://github.com/nathanael-fijalkow/ProgSynth}}.
WikiCoder is written in Python, it leverages a prediction model written in PyTorch.
We favoured simplicity over performance as well as clear separation into independent components as described above.

\paragraph*{Benchmark suite}
Since there are no existing datasets to test the new aspects introduced in Section~\ref{sec:milestones}, we created one ourselves.
The dataset is comprised of 46 tasks in the spirit of the FlashFill dataset~\cite{Gulwani2011}. 
Some of the tasks are inspired or extracted from~\cite{VerbruggenLG21}, they are tagged as such in our code.
Each task requires external knowledge to be solved and is labelled with 3 metadata:
\begin{itemize}
	\item preprocessing for entity extraction: 0 if the inputs are already the sought entities, 1 if a syntactical program is enough to extract them, and 2 otherwise;
	\item complexity of the relationships between entities : 0 when no relation from the knowledge graph is needed, 1 for simple (single edge in the knowledge graph), and 2 for composite;
	\item postprocessing on external knowledge: 0 if the knowledge is used without postprocessing, and 1 otherwise.
\end{itemize} 
This induces 8 categories, we provide at least 4 tasks for each category.

\paragraph*{Knowledge graph}
For simplicity and reproducibility, we use a custom-made small knowledge graph for the experiments. We provide \texttt{SPARQL} queries to construct the knowledge graph.

\paragraph*{Experimental setup}
All our experiments were performed on a consumer laptop (MSI GF65 Thin 9SE) with an intel i7-9750H CPU working up to 2.60GHz on a single thread. The operating system is Manjaro 21.3 with linux kernel 5.17.5-1. The code is written in Python 3.8.10.
The framework used for the \texttt{SPARQL} database is BlazeGraph 2.1.6
The code made available includes all the details to reproduce the experiments, it also includes the exact version of the Python libraries used.

\paragraph*{Prediction Model}
In our prediction model we embed the examples with a one hot encoding.
Since the number of examples can change from one task to another, they are fed into a single layer RNN to build an intermediate representation of fixed shape.
Then it is followed by 4 linear layers with ReLU activation functions, except for the last layer where no activation function is used.
The final tensor is split, and a log softmax operator is applied for each non-terminal of the grammar, thus producing a probabilistic context free grammar in log space.

The prediction model is trained on a dataset of 2500 tasks for 2 epochs with a batch size of 16 tasks.
The tasks were generated with the following process: a program was sampled randomly from a uniform probabilistic context free grammar, then inputs are randomly generated and run on the program.
We used the Adam optimiser with a cross entropy loss to maximise the probability of the solution program being generated from the grammar.

\subsection{Results on the new dataset}
\begin{figure}
	\centering
	\includegraphics[width=0.8\linewidth]{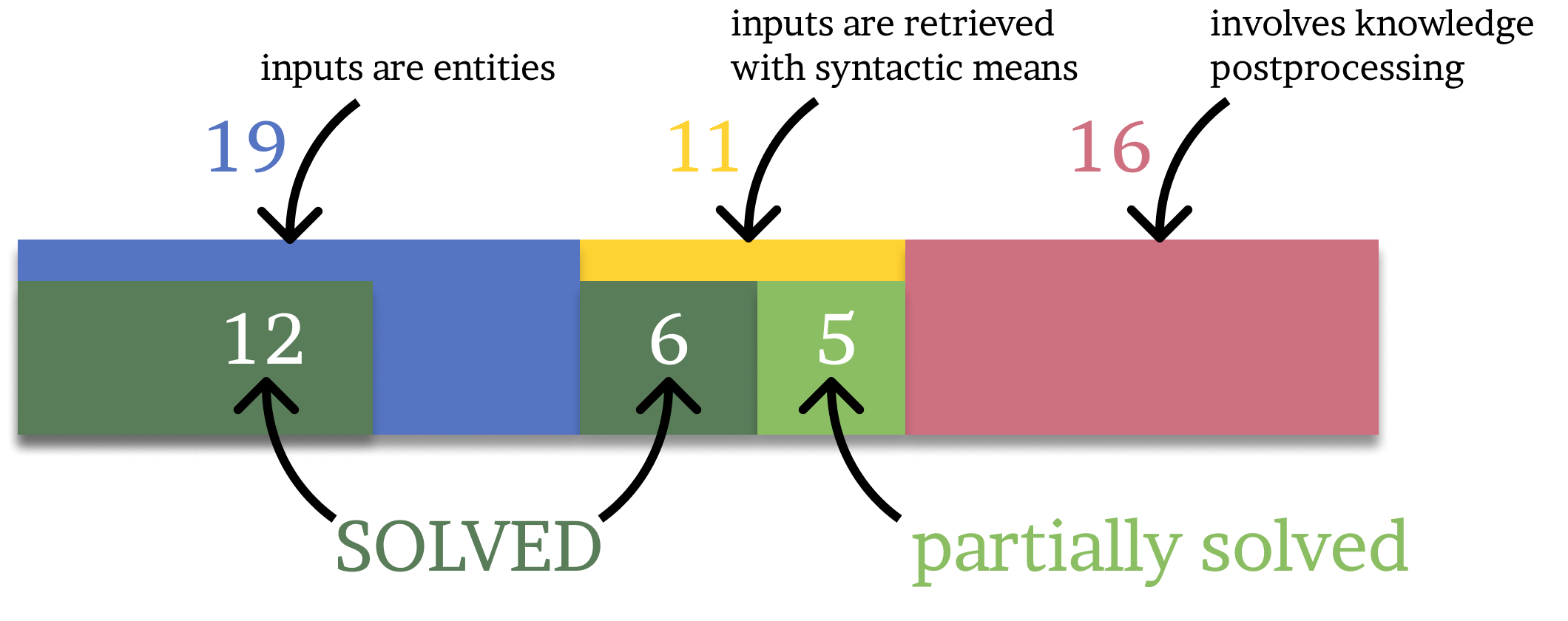}
	\caption{WikiCoder performance on our new dataset}
	\label{fig:results}
\end{figure}
WikiCoder solves 18 out of 46 tasks with a timeout of 60 seconds.
Let us make more sense of this number:
\begin{itemize}
	\item Since our algorithm does not perform postprocessing on knowledge, none of the 16 tasks involving knowledge postprocessing were solved.
	Thus only 30 tasks are within reach of our algorithm.
	\item Among the 30 tasks, for 19 of them the inputs are directly the sought entities.
	WikiCoder solves 12 out of these 19 tasks, most failed tasked are not solvable with our DSL.
	\item In the remaining 11 tasks, the entities can be retrieved using syntactic manipulations.
	WikiCoder solves 6 out of these 11 tasks.
	\item Digging deeper for the last case, the fault lies in all remaining 5 cases with the preprocessing algorithm, which fails to correctly decompose the task and formulate the appropriate \texttt{SPARQL} queries. 
	However, when provided with the right decomposition, WikiCoder solves all 5 tasks.
\end{itemize}
The results can be visualised on Figure~\ref{fig:results}.
The takeaway message is: WikiCoder solves almost all tasks which does not involve knowledge postprocessing, and when it does not the main issue is entities extraction with the preprocessing algorithm.

\subsection{Comparison with Codex and GPT-3}
It is not easy to perform a fair comparison with Codex and GPT-3 as they solve different problems.
There are two ways these models can solve programming by example tasks.
The experiments were performed on OpenAI's beta using the Da Vinci models (the most capable) for GPT-3 and Codex, with default parameters.

\vskip1em
Using Codex: as \textit{code completion}, provided with examples as docstring. This makes the problem very hard for Codex for two reasons: first, it was trained to parse specification in natural language, possibly enriched with examples. Providing only examples is a very partial specification which Codex is not used to. Second, Codex does not use a domain specific language, but rather general purpose programming languages such as Python and C. This implies that the program might be wrong in different ways: incorrect syntax, typing issues, or compilation errors.
For the reasons above, in this infavourable comparison Codex solves only the 5 easiest tasks.

\vskip1em
There are two ways of using GPT-3: \textit{query answering} or \textit{code completion}. 
For query answering, we feed GPT-3 with all but the last examples, and ask it to give the output for the input of the last example:
\begin{minted}{python}
Query:
    f("France") = "I live in France"
    f("Germany") = "I live in Germany"
    f("Poland") = "I live in Poland"
    f("New Zealand") = ?
Output:
    f("New Zealand") = "I live in New Zealand"
\end{minted}
The weakness of this scenario is that GPT-3 does not output a program: it only answers queries. One consequence is that the correctness test is very weak: getting the right answer on a single query does not guarantee that it would on any input. Worse, not outputting a program means that the whole process acts as a black-box, giving up on the advantages of our framework (see related work section) and program synthesis in general.

Using GPT-3/ChatGPT as \textit{code completion}: the prompt is to generate a Python function that satisfies the following examples. This fixes the weakness of not outputting a program with query answering. However, on tasks that require a knowledge graph, the answer is either a succession of if statements or using a dictionnary which is semantically equivalent to the if statements.
In some sense, the system having only partial knowledge, it does not attempt to generalise beyond the given examples.
%From another point of view, it can be seen as GPT-3 seeing the dictionnary as some partial knowledge graph that is left to the user to complete.

GPT-3 performs very well in the query answering setting, solving 31 tasks out of 46. Only 2 tasks were solved by WikiCoder and not by GPT-3, and conversely 11 by GPT-3 and not by WikiCoder. GPT-3 only solves 3 tasks involving knowledge postprocessing: in this sense, WikiCoder and GPT-3 suffer from the same limitations, they both struggle with knowledge postprocessing.

\subsection{Results on FlashFill}
\begin{figure}
	\centering
	\includegraphics[width=0.8\linewidth]{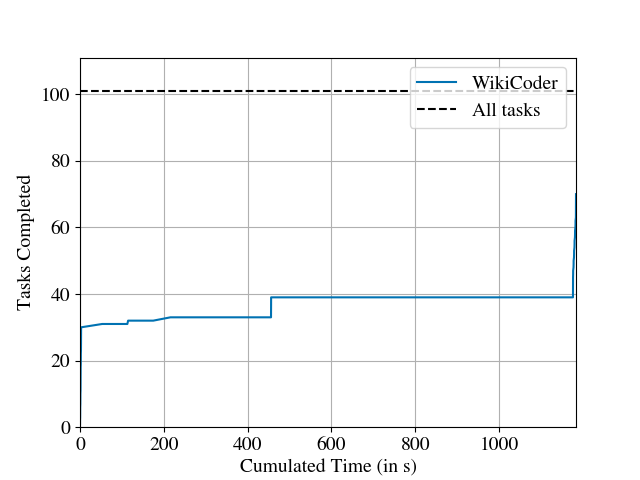}
	\caption{WikiCoder solves 70\% of FlashFill tasks}
	\label{fig:flashfill}
\end{figure}

To show that WikiCoder operates at scale on classical program synthesis tasks, we test it against the classical and established FlashFill dataset.
The results are shown on Figure~\ref{fig:flashfill} (displaying cumulated time), WikiCoder solves 70 out of 101 tasks with a timeout of 60 seconds per task, on par with state of the art general purpose program synthesis tools.
Only 85 tasks can be solved with our DSL.

\subsection{Limitations}
To show the limitations of our approach, we discuss some counterexamples.

\paragraph*{Entities extraction} 
Our preprocessing algorithm looks for the longest shared pattern of length at least 2 in the examples.
Let us consider the following example:
\begin{minted}{python}
f("France") = "I live in France"
f("Germany") = "I live in Germany"
f("Poland") = "I live in Poland"
\end{minted}
The longest pattern is naturally ``I live in '', however another pattern occurs, unexpectedly: ``an'' appears in each country's name. When using our preprocessing algorithm this leads to a wrong sketch.
Adding a fourth example would remove this ambiguity if the country name does not include ``an''.
A related example would be if all examples include year dates in the same millenium, say all of the form ``20XX'': then ``20'' appears in each example, but it should not be separated from XX.

\paragraph*{Knowledge postprocessing} 
None of the tasks involving knowledge postprocessing were solved. Indeed, if the entity to be used is not present in the example, it is very hard to guess which one it is. Natural candidates are entities in the neigbourhood of the starting entity. Our approaches to this challenge have proved inefficient.

\subsection{Examples of Programs}

To show that the programs generated are clear and interpretable by humans, we show a few of them translated into Python equivalents. 
%Note that one function call in the DSL means one function call in Python.
Here is the first example where two relations are needed:
\begin{minted}{python}
// Examples:
    f("France") = "French, capital:Paris"
    f("Germany") = "German, capital:Berlin"
    f("China") = "Chinese, capital:Beijing"
    f("New Zealand") = "New Zealander, capital:Wellington"

// Generated program:
def f(x: str) -> str:
    a = label(follow_edges_from(x, "demonym"))
    b = ", capital:"
    c = label(follow_edges_from(x, "isCapitalOf"))
    return a + b + c
\end{minted}
We present another example where a relation at distance 2 is needed:
\begin{minted}{python}
// Examples:
    f("Paris") = "The phone country code is 33"
    f("Berlin") = "The phone country code is 49"
    f("Detroit") = "The phone country code is 1"
    f("Chihuahua") = "The phone country code is 52"

// Generated program:
def f(x: str) -> str:
    a = "The phone country code is " 
    b = label(follow_edges_from(x, "CityOf", "phoneCode"))
    return a + b
\end{minted}
%We decided to not add other examples of tasks solved since apart from the program synthesis perspective they do not add any information.

%%%%%%%%%%%%%%%%%%%%%%%%%%%
\section{Discussion}
\subsection{Related work}
The closest related work is~\cite{VerbruggenLG21}: 
the tool FlashGPT3 solves knowledge-powered program synthesis with a completely different approach: 
instead of querying a knowledge graph, FlashGPT3 uses a large language model (GPT3~\cite{BrownMRSKDNSSAA20}) to get external knowledge.
FlashGPT3 is shown superior to both program synthesis tools and GPT3 taken in isolation, and achieves impressive results in different domains.
There are three advantages of using a knowledge graph over a large language model:
\begin{itemize}
	\item \textbf{Reliability}: the synthesized program is only as reliable as the knowledge source that it uses.
To illustrate this point, let us consider the task
\begin{minted}{python}
f("Paris") = "France"
f("Berlin") = "Germany"
\end{minted}
GPT3 needs very few examples to predict that \mint{python}|f("Washington") = "United States"| \noindent but it might fail on more exotic inputs.
On the other hand, querying a knowledge graph results in setting \texttt{f = CountryOf} implying that the program will correctly reproduce the knowledge from the graph. 
Although GPT3 has achieved extraordinary results in query answering, it is still more satisfactory to rely on an established and potentially certified source of knowledge such as Wikipedia.
	\item \textbf{Explainability}: the constructed program has good explainability properties: the exact knowledge source it uses can be traced back to the knowledge graph and therefore effectively verified.
	\item \textbf{Adaptability}: knowledge-powered program synthesis can be deployed with any knowledge graph, possibly collecting specialised or private information which a largue language model would not know about.
\end{itemize}

\paragraph*{Other knowledge-powered program synthesis tools}
Most program synthesis tools work at a purely syntactical level. However, some included limited level of semantic capabilities for domain-specific tasks: for instance Transform-data-by-example (TDE) uses functions from code bases and web forms to allow semantic operations in inductive synthesis~\cite{HeCGZNC18},
and APIs were used for data transformations~\cite{BhupatirajuSMK17}.

\paragraph*{Knowledge graphs}
A growing research community focuses on creating, maintaining, and querying knowledge graphs.
The most related problems to our setting are query by example, where the goal is either to construct a query from a set of examples~\cite{JayaramKLYE16,MetzgerSS17}, and entity set expansion, aiming at expanding a small set of examples into a more complete set of entities having common traits~\cite{ZhengSCLW22}.

\subsection{Contributions and Outlook}

We have introduced knowledge-powered program synthesis, extending program synthesis by allowing programs to rely on knowledge graphs.
We described a number of milestones for this exciting and widely unexplored research objective, and proposed a dataset to evaluate the progress in this direction.
We constructed an algorithm and implemented a general-purpose knowledge-powered program synthesis tool WikiCoder that solves tasks previously unsolvable. Our tool can only address about one third of the dataset; we believe that solving the whole dataset would be an important step forward towards deploying program synthesis in the real world.

The most natural continuation of this work is to use very large language models for knowledge-powered program synthesis.
As discussed above, how can we retain the properties of our framework with knowledge graphs: reliability, explainability, and adaptability, while leveraging the power of very large language models?

\bibliographystyle{abbrv}
\bibliography{bib}

\end{document}